%% file: main.tex
\pdfoutput=1

\documentclass[11pt]{article}
\usepackage[]{acl}

\usepackage{times}
\usepackage{latexsym}
\usepackage{hyperref}
\usepackage{multirow}
\usepackage{amsmath}

\usepackage[T1]{fontenc}

\usepackage[utf8]{inputenc}

\usepackage{microtype}

\usepackage{color}
\usepackage{float}
\usepackage{graphicx}
\usepackage{soul}

\title{ETMS@IITKGP at SemEval-2022 Task 10: Structured Sentiment Analysis Using A Generative Approach}

\author{R Raghav\thanks{Equal contribution} \and Adarsh Vemali\footnotemark[1] \and Rajdeep Mukherjee \\
Indian Institute of Technology Kharagpur, India \\ \texttt{\{rraghav5600, adarsh.vemali, rajdeep1989\}@iitkgp.ac.in}}

\begin{document}
\maketitle
\begin{abstract}
Structured Sentiment Analysis (SSA) deals with extracting opinion tuples in a text, where each tuple ($h, e, t, p$) consists of \textit{h}, the holder, who expresses a sentiment polarity \textit{p} towards a target \textit{t} through a sentiment expression \textit{e}. While prior works explore graph-based or sequence labeling-based approaches for the task, we in this paper present a novel unified generative method to solve SSA, a SemEval-2022 shared task. We leverage a BART-based encoder-decoder architecture and suitably modify it to generate, given a sentence, a sequence of opinion tuples. Each generated tuple consists of seven integers respectively representing the indices corresponding to the start and end positions of the holder, target, and expression spans, followed by the sentiment polarity class associated between the target and the sentiment expression. We perform rigorous experiments for both Monolingual and Cross-lingual subtasks, and achieve competitive Sentiment F1 scores on the leaderboard in both settings.

\end{abstract}

\input{Sections/introduction}

\input{Sections/background}
\input{Sections/methodology}

\input{Sections/expt_setup}
\input{Sections/results}

\input{Sections/related_work}
\input{Sections/conclusion}

\bibliography{main, datasets}
\bibliographystyle{acl_natbib}




\end{document}

%% file: Sections/introduction.tex
\section{Introduction}
Structured Sentiment Analysis (SSA) is the task of extracting structured information around sentiment expressions present in text in the form of opinion tuples $O = \{O_1, O_2, ..., O_n\}$, where each opinion tuple $O_i = (h, t, e, p)$ consists of $h$, the \textit{holder} (or \textit{source}, used interchangeably) who expresses a sentiment polarity $p$ towards a \textit{target} (or \textit{aspect}) $t$ using an opinion or sentiment expression $e$ \cite{barnes2021structured}. Prior works \cite{liu2012sentiment, Peng2020KnowingWH} have highlighted the importance of addressing sentiment analysis as a structured prediction problem in order to capture the complete information around various opinions expressed in the text. The task of SSA thus expects to exploit the pair-wise interactions between the members of the same opinion tuple during the extraction process.

With the exponential growth of online marketplaces and user-generated content therein, SSA or near similar tasks of aspect-sentiment-opinion triplet extraction \cite{Peng2020KnowingWH, mukherjee-etal-2021-paste, yan2021unified}, and aspect-category-sentiment-opinion quad extraction \cite{cai-etal-2021-aspect}, the newest additions under the broader umbrella of aspect-based sentiment analysis (ABSA) \cite{pontiki-etal-2015-semeval, pontiki-etal-2016-semeval} have become more important than ever \cite{mukherjee_ecir21}. In the face of ever-expanding choices, it becomes a challenging necessity to take educated explainable decisions from past user reviews. SSA guides the learning in the proper direction by facilitating an automated way to focus on major sentiment or opinion indicators. As a result, the task has wide applications in various market segments, such as e-commerce, food delivery, healthcare, ride sharing, travel and hospitality, to name a few.

\begin{table*}[!ht]
    \centering
    \begin{tabular}{|l|c|c|c|}
    \hline
        \multicolumn{1}{|c|}{Dataset Name} & Language & \% Null (Train, Dev) & Size (Train, Dev, Test) \\ \hline
        NoReC\_fine \cite{ovrelid-etal-2020-fine} & Norwegian & (47.24\%, 46.37\%) & (8634, 1531, 1272) \\ 
        MultiBooked\_eu \cite{barnes-etal-2018-multibooked} & Basque & (15.43\%, 21.05\%) & (1064, 152, 305) \\ 
        MultiBooked\_ca \cite{barnes-etal-2018-multibooked} & Catalan & (14.65\%, 16.17\%) & (1174, 168, 336) \\ 
        OpeNER\_es \cite{Agerri2013} & Spanish & (12.93\%, 15.53\%) & (1438, 206, 410) \\ 
        OpeNER\_en \cite{Agerri2013} & English & (19.72\%, 20.48\%) & (1744, 249, 499) \\
        MPQA \cite{Wiebe2005b} & English & (77.92\%, 79.18\%) & (5873, 2063, 2112) \\ 
        Darmstadt\_unis \cite{toprak-etal-2010-sentence} & English & (69.77\%, 64.66\%) & (2253, 232, 318) \\ 
    \hline
    \end{tabular}
    \caption{Dataset Statistics}
    \label{table:data-structure}
\end{table*}

\begin{figure*}[!ht]
\centering
\includegraphics[width=0.9\textwidth]{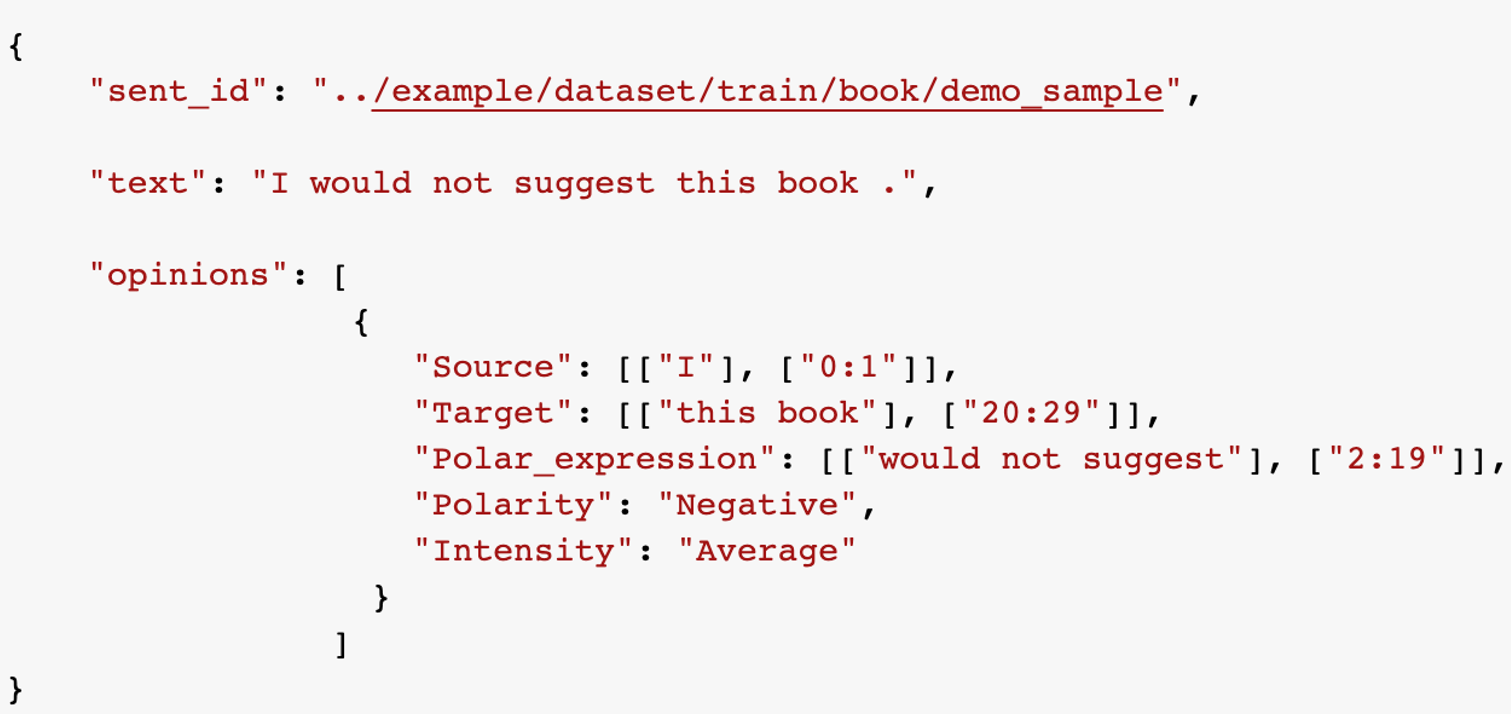}
\caption{Annotation Format. The value of "opinions" in the JSON is a list of opinion tuples present in the "text". Each item in the list is a dictionary, with keys being the tuple elements, and the values corresponding to their representation in the text. \textit{Source, Target} and \textit{Polar\_expression} are annotated with the the actual word spans appearing in the text along and their respective character indices. \textit{Polarity} represents the sentiment expressed in the tuple, and \textit{Intensity} represents its strength.}
\label{image:json_example}
\end{figure*}

Previous efforts on SSA have primarily focused on two approaches: sequence labeling-based \cite{he-etal-2019-interactive}, and graph-based \cite{barnes-etal-2021-structured}. The former tries to first predict the presence/absence of targets and expressions in the text by sequentially labeling each text token using BIOES\footnote{BIOES is a tagging scheme commonly used for sequence labeling tasks. B, I, E, O, and S respectively denote the \textit{begin, inside, outside, end}, and \textit{single} tags corresponding to an entity.} tags, before modeling their interaction to predict the sentiment polarity. The latter models the task as a dependency graph parsing problem, where the sentiment expression is considered as the root node, and the other elements are connected via arcs that represent their relationships. Different from these, we present a novel generative approach to solve SSA. More specifically, we take motivation from a unified generative framework recently proposed by \citet{yan2021unified} to solve several ABSA tasks. We suitable modify their BART-based encoder-decoder architecture to adapt it for SSA. Given a sentence, the model is trained to generate a sequence of tuples, each consisting of seven integer outputs corresponding to the start and end indices of the holder, target and sentiment expression spans appearing in the text, and finally the polarity class. An example is shown and described in Figure \ref{image:model_architecture}.

We participate in the SemEval 2022 Task 10: Structured Sentiment Analysis \cite{barnes-etal-2022-semeval} hosted on \href{https://competitions.codalab.org/competitions/33556}{CodaLab}. In order to demonstrate the efficacy of our proposed solution, we attempt both the \textit{monolingual} and \textit{cross-lingual} subtasks and achieve competitive performance on the leaderboard in both settings. As part of the (sub)tasks, we testify our approach on multiple datasets spanning across five different languages - \textit{English} (Darmstadt\_unis, OpeNER\_en, MPQA), Basque (MultiBooked\_eu), \textit{Catalan} (MultiBooked\_ca), \textit{Norwegian} (NoReC\_fine) and \textit{Spanish} (OpeNER\_es). A summary of dataset statistics is reported in Table \ref{table:data-structure}. While the evaluation scripts were made available to us by the task organizers to analyze our performance, the final leaderboard scores were obtained on a hidden test set.

%% file: Sections/background.tex
\section{Task Overview}
\subsection{Task Definition}
SSA aims to predict all the structured sentiment graphs present in a given text. A graph is formally represented by opinion tuples $O = O_1, O_2, ... , O_n$, where each opinion tuple $O_i$ consists of a quadruple of the holder $h$, the target $t$, the sentiment expression $e$, and the sentiment polarity $p$.

\begin{figure*}[!ht]
\includegraphics[width=\textwidth]{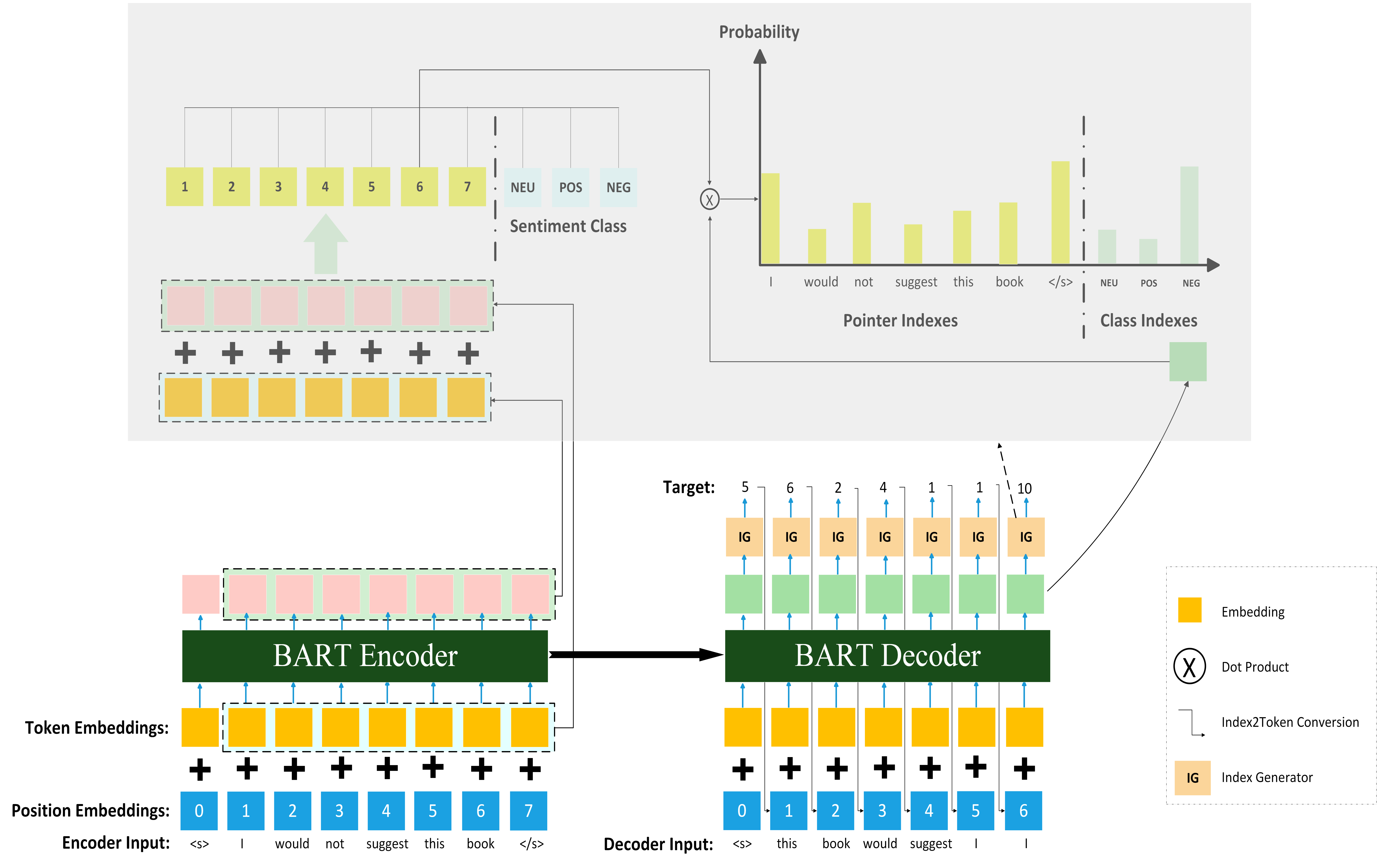}
\caption{Model Architecture. The above figure shows an example where the input is "<s> I would not suggest this book </s>", and the corresponding output is "5, 6, 2, 4, 1, 1, 10" (Only partial decoder sequence is shown. Here, 7 (</s>) should be the next generation index). The "Index2Token Conversion" module converts the pointer indices back to the corresponding tokens in the source text, and the class index to the corresponding sentiment polarity.
}
\label{image:model_architecture}
\end{figure*}

\subsection{Datasets}
As summarized in Table \ref{table:data-structure}, we are provided with a total of 7 datasets, as part of the shared task, spanning across 5 different languages. Each dataset is a collection of sentences, along with their corresponding annotated opinion tuples, each consisting of (\textit{Source, Target, Polar Expression}, and \textit{Polarity}). While the \textit{Intensity} of the expressed sentiment is also provided as part of the annotations, intensity classification/regression is not included as part of the task. An example is shown in Figure \ref{image:json_example}. All the data is provided through \href{https://competitions.codalab.org/competitions/33556}{CodaLab}, and \href{https://github.com/jerbarnes/semeval22_structured_sentiment}{GitHub}.

%% file: Sections/methodology.tex
\section{Methodology}

\subsection{Task Formulation}
\label{section: Task Formulation}
We take a generative approach to formulate SSA as a structured prediction problem. We note here that predicting the holder (source), target (aspect), and polar expression spans correspond to extraction tasks, whereas sentiment polarity prediction is a classification task. Following \cite{yan2021unified}, we model both these tasks in a unified generative framework by representing span entities with their start and end pointer indices corresponding to the text, and sentiment polarity with a class index. 

We denote the holder, target, polar expression and sentiment polarity as \textit{h}, \textit{t}, \textit{pe}, and \textit{sp} respectively. The start and end index of each term is represented using superscripts $^s$ and $^e$. For an input $X = [x_1,...,x_n]$, where $x_i$ is the $i^{th}$ word in the text, the target $Y = [t_i^s, t_i^e, pe_i^s, pe_i^e, h_i^s, h_i^e, sp_i, ...]$ is defined as a sequence of tuples, each consisting of seven indices corresponding to an opinion tuple $(h, t, pe, sp)$. An example sentence along with its target sequence are shown in Figure \ref{image:model_architecture}.

\subsection{System Overview}
Our model, as shown in Figure \ref{image:model_architecture}, consists of an encoder-decoder architecture with BART \cite{lewis2019bart} as its backbone. Given an input $X = [x_1, ..., x_n]$, the model is trained to produce an output $Y = [y_1, ..., y_m]$ (with $y_0$ representing the start-of-sequence token, $<s>$). The probability distribution is modeled as:
\begin{equation}
    P(Y|X) = \prod_{t=1}^{m} P_t
\end{equation}

Here $P_t = P(y_t|X,Y_{<t})$ represents the index probability distribution for the $t^{th}$ time step.

\subsubsection{Encoder}
BART comprises of a bi-directional encoder. We denote the encoded vector of the input sentence $X$ as $H^e$. For the sake of simplicity, we ignore the start-of-sequence token $(<s>)$ in the equations.
\begin{equation}
    H^e = BARTEncoder([x_1, ..., x_n])
\end{equation}
Here $H^e \in R^{n \times d}$, with d as the hidden dimension.

\begin{table}[H]
    \centering
    \begin{tabular}{|l|c|}
    \hline
        \multicolumn{1}{|c|}{Dataset} & \% Null Instances \\ \hline
        NoReC\_fine & 47.24\% (As in the dataset) \\
        MultiBooked\_eu & 10\% \\ 
        MultiBooked\_ca & 14.65 \% (As in the dataset) \\
        OpeNER\_es & 4\% \\
        OpeNER\_en & 14\% \\
        MPQA & 50\% \\ 
        Darmstadt\_unis & 69.77 \% (As in the dataset) \\ \hline
    \end{tabular}
    \caption{\% null instances in processed train sets.}
    \label{table:per_null}
\end{table}

\subsubsection{Index2Token Conversion}\label{subsubsec:conversion}
Since the entity spans are decoded as corresponding start and end indices, and the sentiment polarity is decoded as corresponding class index, the indices need to be converted back to tokens before the BART Decoder can use them along with the encoder hidden state $H^e$ for generating the next token (index) in the $t^{th}$ time step. For each $y_t \in Y_{<t}$, we therefore use the following conversion strategy:
\begin{equation}
    \hat{y}_{t} =
    \begin{cases}
      X_{y_t} & \text{if $y_t$ is a pointer index,}\\
      Pol_{y_t-n} & \text{if $y_t$ is a class index}
    \end{cases}
\end{equation}
where $Pol = [p_1, p_2, p_3]$ is the list of polarity classes. In our implementation, $y_t \in [1, n + 3]$. The first sentence token $x_1$ has the pointer index 1.

\subsubsection{Decoder}
Our BART decoder now uses $H^e$ and the converted decoder outputs $\hat{Y}_{<t}$ to obtain the $t^{th}$ decoder hidden state:
\begin{equation}
    H^d_t = BARTDecoder(H^e, \hat{Y}_{<t}) 
\end{equation}
where $H^d_t \in R^{d}$. Finally, $H^d_t$ is used to predict the token probability distribution $P_t$. We request our readers to refer to \cite{yan2021unified} for additional details.

\subsubsection{Training and Workflow}
Teacher forcing with negative log likelihood as the loss function is used to train the model. During inference, beam search is used to generate the target sequence $Y$ in an auto-regressive manner. Finally, the generated sequence is translated back into the phrase spans and sentiment polarity. As shown in Figure \ref{image:model_architecture}, we now illustrate the working of our proposed method using an example sentence: \\\textit{I would not suggest this book .}

\begin{enumerate}
    \item The input <s> I would not suggest this book </s> is sent as input to the BART Encoder. As specified in Section \ref{subsubsec:conversion}, the word "I" is mapped to position index 1. Accordingly, </s> is mapped to index 7. Thereafter, each sentiment polarity is assigned a class index in sequence. In this case, the polarity values neutral, positive, and negative, are assigned class indices 8, 9 and 10 respectively.
    \item The BART Decoder is trained to generate a sequence of indices till the end-of-sequence index (here 7) is generated. Corresponding to each opinion tuple, the decoder respectively predicts the start and end word indices for the target, polar expression, and source and finally, the polarity class index. 
    \item In our case, the expected target sequence is 5, 6, 2, 4, 1, 1, 10, 7. Here, (5, 6) represents the \textit{target} phrase "this book", (2, 4) represents the \textit{polarity expression} phrase "would not suggest", (1, 1) represents the \textit{holder} phrase "I", and 10 represents the \textit{negative} polarity class.
    \item During inference, a decoding algorithm is employed making use of the \textit{Index2Token Conversion} module to respectively convert the indices back to the text tokens and polarities before presenting to the end user.
    

\end{enumerate}

%% file: Sections/expt_setup.tex
\section{Experiments}

\subsection{Data Preprocessing}

\begin{table*}[!ht]
    \centering
    \small
    \begin{tabular}{|c|c|c|c|c|c|}
    \hline
        Dataset & \% null & Batch Size & LR & F1 Score & Best Epoch \\ \hline

        NoReC\_fine & 10 & 16 & 1E-04 & 0.320 & 35 \\ 
        NoReC\_fine & 10 & 16 & 2E-05 & 0.310 & 8 \\ 
        NoReC\_fine & 10 & 16 & 5E-05 & 0.323 & 11 \\ \hline
        \hline
        OpeNER\_es & 10 & 8 & 1E-04 & 0.351 & 33 \\ 
        OpeNER\_es & 10 & 8 & 2E-05 & 0.482 & 36 \\ 
        OpeNER\_es & 10 & 8 & 5E-05 & 0.566 & 33 \\ \hline
        \hline
        OpeNER\_en & 10 & 8 & 1E-04 & 0.674 & 43 \\ 
        OpeNER\_en & 10 & 8 & 2E-05 & 0.661 & 32 \\ 
        OpeNER\_en & 10 & 8 & 5E-05 & 0.675 & 7 \\ \hline
        \hline
        Darmstadt\_unis & 10 & 16 & 1E-04 & 0.259 & 30 \\ 
        Darmstadt\_unis & 10 & 16 & 2E-05 & 0.276 & 34 \\ 
        Darmstadt\_unis & 10 & 16 & 5E-05 & 0.289 & 19 \\ \hline
    \end{tabular}
    \caption{Learning Rate Tuning}
    \label{table:learning rate tuning}
\end{table*}

\begin{table*}[!ht]
    \centering
    \small
    \begin{tabular}{|c|c|c|c|c|c|}
    \hline
        Dataset & \% null & Batch Size & LR & F1 Score & Best Epoch \\ \hline
        
        NoReC\_fine & 15 & 8 & 5E-05 & 0.317 & 8 \\ 
        NoReC\_fine & 30 & 16 & 5E-05 & 0.295 & 17 \\ 
        NoReC\_fine & As in Dataset & 16 & 5E-05 & 0.357 & 28 \\ \hline
        \hline
        
        MultiBooked\_eu & 5 & 8 & 5E-05 & 0.422 & 20 \\ 
        MultiBooked\_eu & 10 & 8 & 5E-05 & 0.427 & 21 \\ 
        MultiBooked\_eu & As in Dataset & 8 & 5E-05 & 0.401 & 19 \\ \hline
        \hline
        
        MultiBooked\_ca & 5 & 8 & 5E-05 & 0.552 & 35 \\ 
        MultiBooked\_ca & 10 & 8 & 5E-05 & 0.545 & 32 \\ 
        MultiBooked\_ca & As in Dataset & 8 & 5E-05 & 0.556 & 46 \\ \hline
        \hline
        
        OpeNER\_es & 4 & 8 & 5E-05 & 0.572 & 31 \\ 
        OpeNER\_es & 8 & 8 & 5E-05 & 0.57 & 4 \\ 
        OpeNER\_es & As in Dataset & 8 & 5E-05 & 0.571 & 25 \\ \hline
        \hline
        
        OpeNER\_en & 7 & 16 & 5E-05 & 0.677 & 24 \\ 
        OpeNER\_en & 14 & 16 & 5E-05 & 0.678 & 25 \\ 
        OpeNER\_en & As in Dataset & 16 & 5E-05 & 0.674 & 39 \\ \hline
        \hline
        
        MPQA & 25 & 8 & 5E-05 & 0.355 & 20 \\ 
        MPQA & 50 & 16 & 5E-05 & 0.366 & 44 \\ 
        MPQA & As in Dataset & 16 & 5E-05 & 0.359 & 42 \\ \hline
        \hline
        
        Darmstadt\_unis & 25 & 16 & 5E-05 & 0.268 & 42 \\ 
        Darmstadt\_unis & 45 & 16 & 5E-05 & 0.277 & 33 \\ 
        Darmstadt\_unis & As in Dataset & 16 & 5E-05 & 0.312 & 36 \\ \hline

    \end{tabular}
    \caption{Null Parameter Tuning for each dataset}
    \label{table:null parameter tuning}
\end{table*}

In addition to fully annotated sentences, we observe the following two kinds of instances in all the datasets: (a) no opinion tuples at all - hereby referred to as the null examples, and (b) few empty entities in a single opinion tuple. We note here that our tuple representation scheme expects position indices of words appearing in the sentence. In order to accommodate for the above mentioned cases, we add a string \textit{None} in the beginning of every sentence. This helps us to map the missing entities (e.g. holder or aspect in an opinion tuple) to a phrase present in the sentence. 

After rigorous experiments, we set an optimal threshold for the proportion of null examples to be used for training in each of the datasets as reported in Table ~\ref{table:per_null}. During training, we found that limiting the proportion of null examples to the reported values significantly helped us in achieving the best performance on the respective datasets across monolingual and cross-lingual settings.

\subsection{Experimental Setup}
For the \textit{monolingual} setting, we used the train, validation, and test splits of the same datasets. While experimenting on the \textit{English} datasets (Darmstadt\_unis, MPQA, and OpeNER\_en), we use BART-base\footnote{\url{https://huggingface.co/facebook/bart-base}} as the backbone. For the \textit{Non-English} datasets (OpeNER\_es, Multibooked\_eu, Multibooked\_ca, and NoReC\_fine), we use BART-large-MNLI \footnote{\url{https://huggingface.co/facebook/bart-large-mnli}} as the backbone. In the \textit{cross-lingual} setting, we trained our models using the combined training data from all \textit{English} datasets and evaluated them on the test sets of respective \textit{Non-English} datasets (NoReC\_fine not included as part of this setting). Here, we used BART-large-MNLI as the backbone for all our cross-lingual experiments.


Although predicting the intensities of sentiment polarities was not included as part of the shared task, we hypothesized that additionally learning the intensity prediction task would help the model in predicting the other entities (h, t, e, p) better in a multi-task setting. We performed additional experiments to verify our hypothesis. However, we observed little to no difference in the final results. Accordingly, we excluded intensity prediction from further consideration while performing our final experiments. We make our code repository publicly available at \url{https://github.com/Sherlock-Jerry/SSA-SemEval}.

\subsection{Hyperparameter Tuning}
\begin{table}[!ht]
    \centering
    \begin{tabular}{|l|c|c|}
    \hline
        \multicolumn{1}{|c|}{Parameters} & English & Non-English \\ \hline
        Batch Size & 16 & 8 \\ 
        Learning Rate & 5E-05 & 5E-05 \\ 
        Epochs & 50 & 50 \\ 
        BART Model & Base & Large-MNLI \\ \hline
        \% Null & \multicolumn{2}{c|}{Varies with Dataset}  \\\hline
    \end{tabular}
    \caption{Final set of hyperparameters}
    \label{table:hyperparameters}
\end{table}

We train all our models on Tesla P100-PCIE 16GB GPU. We perform extensive tuning experiments to obtain the optimal set of hyperparameters. To determine the optimal learning rate, we ran monolingual experiments on two English and Non-English datasets respectively, as elucidated in Table \ref{table:learning rate tuning}. Based on our observations, we fixed a common learning rate of $5e-5$ for our final experiments across both the settings, monolingual as well as cross-lingual. For obtaining the optimal proportion of null instances to be used for training the final models, we perform three iterations of monolingual experiments on each dataset, each time with a different proportion of null instances used for training the models, as reported in Table \ref{table:null parameter tuning}. The final null thresholds are reported in Table \ref{table:per_null}. Table \ref{table:hyperparameters} summarizes the set of hyperparameters used for reporting our final results for both the subtasks. For all our experiments, the model selected according to the best F1 score on the validation data was used to evaluate on the test data.


\subsection{Evaluation Metrics}
\textit{Sentiment Graph F1} (SG-F1) is used to evaluate the models. \textit{True Positive} is defined as an exact match (including polarity) at graph-level, weighted by the token-level overlap between the gold and predicted spans for holder, target, and polar expression, averaged across all three spans. \textit{Precision} is calculated by weighting the number of correctly predicted tokens divided by the total number of predicted tokens. \textit{Recall} is calculated by dividing the number of correctly predicted tokens by the number of gold tokens, thereby allowing for empty holders and targets which exist in the gold standard.

%% file: Sections/results.tex
\section{Results}

\begin{table}[!t]
    \centering
    \resizebox{.95\columnwidth}{!}
    {
        \begin{tabular}{|c|c|c|}
            \hline
            Dataset & Ours & \citet{barnes-etal-2021-structured} \\ 
            \hline
            NoReC\_fine & 0.351 & 0.312 \\
            MultiBooked\_eu & 0.438 & 0.547 \\ 
            MultiBooked\_ca & 0.508 & 0.568 \\
            OpeNER\_es & 0.544 & Not available \\ 
            OpeNER\_en & 0.626 & Not available \\
            MPQA & 0.327 & 0.188 \\ 
            Darmstadt\_unis & 0.330 & 0.265 \\
            \hline
        \end{tabular}
    }
    \caption{Monolingual SubTask: Test Set SG-F1 Scores.}
    \label{table:monolingual-results}
\end{table}

\subsection{Monolingual Subtask}
We report the results for our monolingual experiments in Table \ref{table:monolingual-results} and compare them with the existing state-of-the-art (SOTA) results reported in \citet{barnes-etal-2021-structured}. Amongst the given datasets, our model performs the best on OpeNER\_en and OpeNER\_es, and has a relatively poor performance on MPQA, Darmstadt\_unis and NoReC\_fine. Despite this, we comfortably outperform the existing SOTA on these datasets. On the public leaderboard hosted on \href{https://competitions.codalab.org/competitions/33556}{CodaLab}, we achieved $18^{th}$ rank out of 32 entries for this task.

\subsection{Crosslingual Subtask}
We report the results for our crosslingual experiments in Table \ref{table:crosslingual-results}. In this paradigm, we used all the English datasets for training our model, and tested our best trained models on the test sets of the respective Non-English datasets.

\begin{table}[H]
    \centering
    \begin{tabular}{|c|c|} 
        \hline
        Dataset & SG-F1 \\ 
        \hline
        EN-EU & \multirow{2}{*}{0.431} \\
        (MultiBooked\_eu) & \\ 
        \hline
        EN-CA & \multirow{2}{*}{0.506} \\
        (MultiBooked\_ca) & \\ 
        \hline
        EN-ES & \multirow{2}{*}{0.542} \\
        (OpeNER\_es) & \\ 
        \hline
    \end{tabular}
    \caption{Corsslingual SubTask: Test Set SG-F1 Scores.}
    \label{table:crosslingual-results}
\end{table}


Here, we achieved $11^{th}$ rank out of 32 entries.

\begin{table*}
    \centering
    \resizebox{.8\textwidth}{!}
    {
    \begin{tabular}{|l|c|l|l|c|}
        \hline
        Label Type & \hspace{0.5cm} Source \hspace{0.5cm} & \multicolumn{1}{|c|}{Target} & \multicolumn{1}{|c|}{Polar Expression} & Polarity \\ 
        \hline
        Gold & - & The size of room & reasonable & Positive \\
        Prediction & - & The size & reasonable & Positive \\ 
        \hline
        Gold & - & walls & in very poor conditions & Negative \\
        Prediction & - & walls & very poor & Negative \\ 
        \hline
        Gold & - & floor & in very poor conditions & Negative \\
        Prediction & - & floor & in very poor conditions & Negative \\ 
        \hline
        Gold & - & ceiling & in very poor conditions & Negative \\
        Prediction & - & ceiling & in very poor conditions & Negative \\
        \hline 
    \end{tabular}
    }
    \caption{Ground truth opinion tuples and model predictions for the sentence: "The size of room is reasonable , but floor , walls and ceiling are in very poor conditions".}
    \label{table:example_sentence}
\end{table*}

\subsection{Qualitative Analysis}
\begin{figure}
    \centering
    \includegraphics[width=\columnwidth]{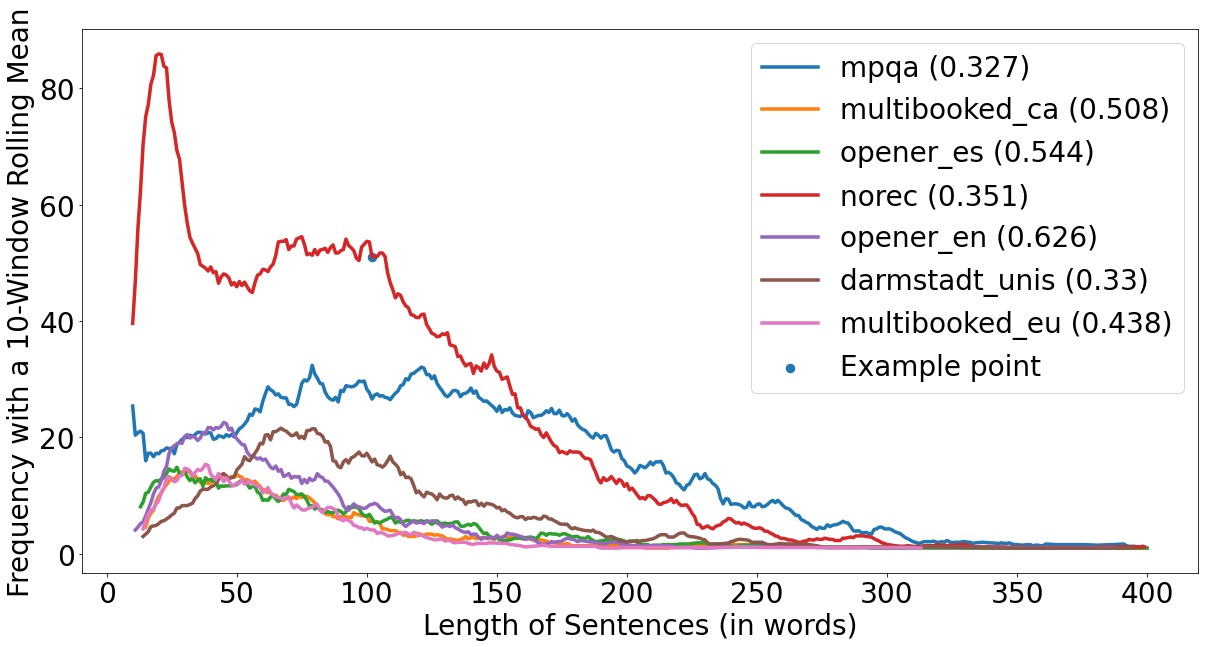}
    \caption{Comparing the distribution of test sentence lengths with best-obtained SG-F1 scores for each dataset. The "Example point" shows that there 51 sentences in the test set for NoRec\_fine with length 100.}
    \label{image:sentence_len_hist}
\end{figure}

\begin{itemize}
    \item In the monolingual subtask, we observed that our model performs poorly on the datasets with large proportions of empty opinion tuples (null instances). As can be confirmed from Table \ref{table:per_null}, MPQA, Darmstadt\_unis, and NoRec\_fine have high empty tuple proportion as against OpeNER\_en, and OpeNER\_es with low proportion of null instances.
    \item We observed that for datasets having lengthy sentences, our model performs relatively poor. A comparison between the distribution of test sentence lengths and the Sentiment Graph F1 scores for each dataset is shown in Figure \ref{image:sentence_len_hist}.
    \item We also observed annotation errors in the datasets. For instance, given the test sentence "So wonderful to see people go to work smiling and leave work still smiling and happy.", our trained generative model correctly predicts an opinion tuple with "see people go to work" as the target, and "So wonderful" as the opinion expression with a "Positive" sentiment. However, no ground truth opinion tuples are associated with the sentence.
    \item As reported in Table \ref{table:example_sentence}, we found a few instances where our model correctly predicts the necessary entities; but due to ambiguity in labelling (even at human level), we saw a mismatch. Here, our model predicts "The size" as the \textit{target} whereas the gold standard expects "The size of room". Similar is the case with "in very poor conditions" (gold standard) versus the predicted phrase "very poor".
\end{itemize}

%% file: Sections/related_work.tex
\section{Related Work}
Previous efforts on SSA have primarily focused on two approaches: sequence labeling-based \cite{he-etal-2019-interactive}, and graph-based \cite{barnes-etal-2021-structured}. The corresponding scores for both these approaches are considered as baselines by the task organizers.

\subsection{Sequence Labelling}
In this approach, \cite{he-etal-2019-interactive} propose a pipeline of sequence labelling and relation classification tasks. More specifically, three different sequence labellers based on BIOES tags are trained to predict the three span-based opinion entities, i.e. the holder, the target, and the polar expression. Finally, their relationship is exploited using a separate classification layer on top to predict the connecting sentiment polarity. However, such an approach inherently suffers from error propagation between the steps. Also, the inter-dependency especially between the target and the polar expression is not captured when the spans are predicted in isolation.
 
\subsection{Dependency Graph Parsing}
\begin{figure}[!ht]
\includegraphics[width=\columnwidth]{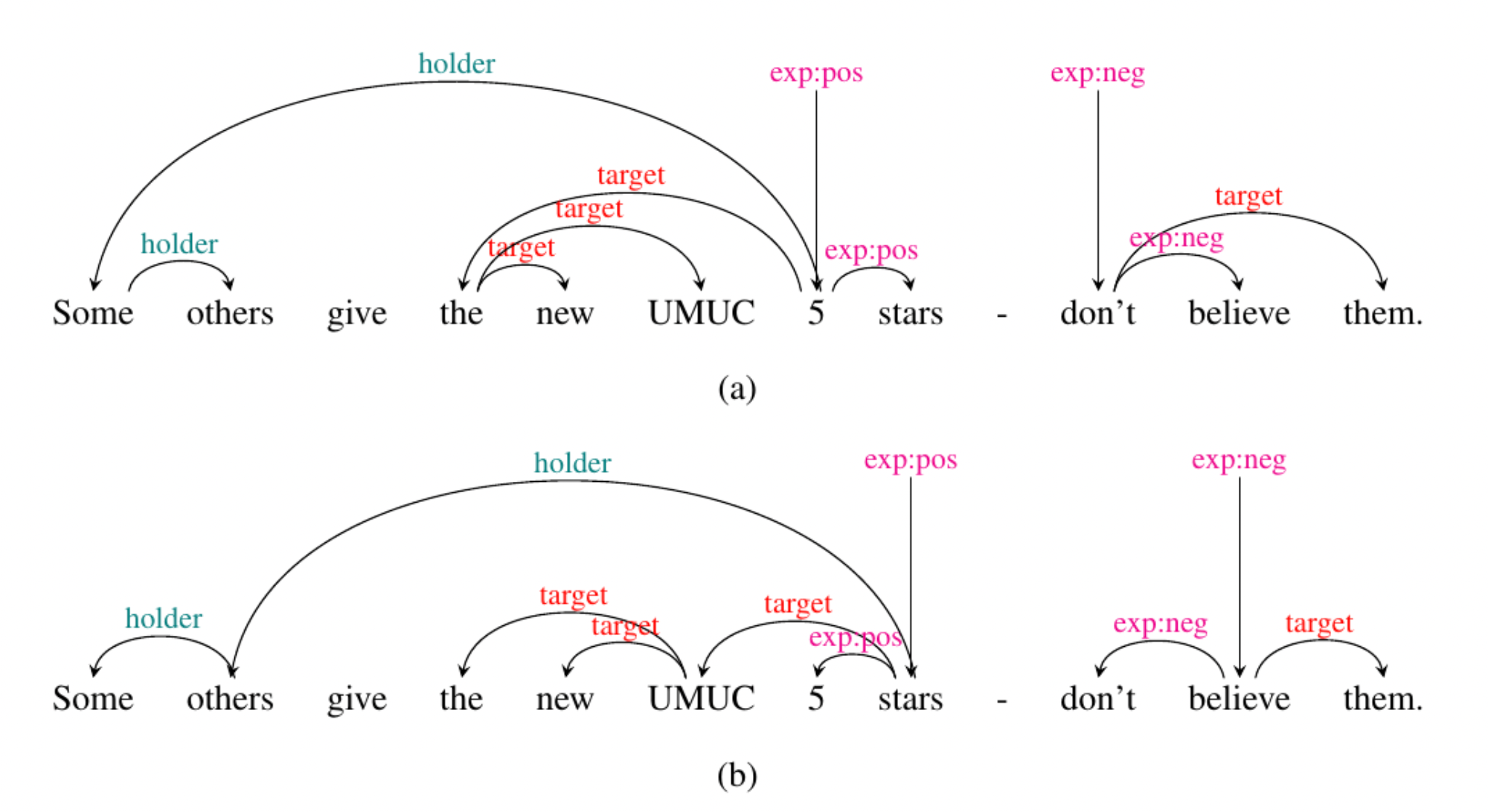}
\caption{Dependency Graph Parsing}
\label{image:dgp_example}
\end{figure}

\cite{barnes-etal-2021-structured} have treated this task as a bilexical dependency graph prediction problem. They present two different versions of their proposed approach - (a) head-first and (b) head-final, as shown in Figure \ref{image:dgp_example}.

In both cases, the sentiment expression is considered as the root node, and the other elements are connected via arcs that represent their relationships. This approach builds upon the \href{https://openreview.net/forum?id=Hk95PK9le}{Dozat and Manning parser}, implemented in \cite{kurtz-etal-2020-end}.

%% file: Sections/conclusion.tex
\section{Conclusion}
Different from prior methods, we in this work present a novel generative approach to tackle the task of Structured Sentiment Analysis. We formulate the task as a structured prediction problem. Our BART-based encoder-decoder architecture is trained to predict a sequence of indices corresponding to each opinion tuple present in the text. The generated indices suitably represent the holder, target, and polar expression spans by their start and end token positions, and the sentiment polarity by its corresponding class. As part of SemEval 2022 Task 10, we participate in both monolingual and crosslingual subtasks, and achieve competitive performance on the leaderboard for both settings. In future, we would like to explore paraphrasing-based generative methods for the task.